\documentclass[10pt,twocolumn,letterpaper]{article}

\usepackage{cvpr}
\usepackage{times}
\usepackage{epsfig}
\usepackage{graphicx}
\usepackage{amsmath}
\usepackage{amssymb}
\usepackage{multirow}
\usepackage{booktabs}
\usepackage{float}
\usepackage{subfig}

\usepackage[pagebackref=true,breaklinks=true,letterpaper=true,colorlinks,bookmarks=false]{hyperref}

\cvprfinalcopy 

\begin{document}

\title{Generative Adversarial Network Architectures For Image Synthesis Using Capsule Networks}

\author{Yash Upadhyay\\
University of Minnesota, Twin Cities\\
Minneapolis, MN, 55414\\
{\tt\small upadh034@umn.edu}
\and
Paul Schrater\\
University of Minnesota, Twin Cities\\
Minneapolis, MN, 55414\\
{\tt\small schrater@umn.edu}
}

\maketitle

\begin{abstract}
In this paper, we propose Generative Adversarial Network (GAN) architectures that use Capsule Networks for image-synthesis. Based on the principal of positional-equivariance of features, Capsule Network's ability to encode spatial relationships between the features of the image helps it become a more powerful critic in comparison to Convolutional Neural Networks (CNNs) used in current architectures for image synthesis. Our proposed GAN architectures learn the data manifold much faster and therefore, synthesize visually accurate images in significantly lesser number of training samples and training epochs in comparison to GANs and its variants that use CNNs. Apart from analyzing the quantitative results corresponding the images generated by different architectures, we also explore the reasons for the lower coverage and diversity explored by the GAN architectures that use CNN critics. 
   
\end{abstract}

\section{Introduction}

Generative Adversarial Networks \cite{goodfellow2014generative} are finding popular applications as generative models in diverse scenarios. One of the biggest advantages of GANs is that they can be trained completely using back-propagation. GANs utilize two adversary multi-perceptron networks (generator and discriminator) that play a minimax game where the generator tries to learn the probability distribution $p_g$ over a dataset $x$. Noise variables $p_z(z)$ serve as an input to the mapping function , the generator, $G(z, \theta_g)$ with parameters $\theta _g$. $D(x,\theta_d)$ represents the discriminator with parameters $\theta _d$ and $D(x)$  represents the probability that the input $x$ came from the dataset rather than $p_g$. Following equation describes the minimax game being played by the adversaries,
\begin{equation}
\begin{aligned}
\min_{G} \max_{D} \mathbb{E}_{x\sim p_{data}(x)}[log(D(x)] + \\
\mathbb{E}_{z\sim p_{z}(z)}[log(1 - D(G(z)))]    
\end{aligned}
\end{equation}
 \cite{arjovsky2017wasserstein} discussed how KL-divergence, as the proposed loss function in \cite{goodfellow2014generative}, can lead to uninformative gradients over low-dimensional manifolds where the intersection between the real and generated data can be very small. Therefore, they introduced Wasserstein GAN (WGAN) that used Earth-Mover's (Wasserstein-1) distance as a loss metric, which is continuous and with near-linear gradients provides a healthy convergence of the generator. 
However, \cite{gulrajani2017improved} demonstrated how the use of a gradient clipping used to enforce 1-Lipschitz continuity was a naive approach and therefore led to very strong regularization of the critic, which ultimately led to underfitting of the model. Therefore, they introduced Gradient Penalty, $\lambda \mathbb{E}_{\hat{x} \sim \mathbb{P}_{\hat{x}}} [(||\nabla_{\hat{x}} D(\hat{x})||_2 - 1)^2 ]$ to the original Wasserstein critic loss which helps bypass the regularization via gradient clipping. The following shows the new loss function for the critic to minimize,
\begin{equation} \label{GPeqn}
\begin{aligned}
    L = \mathbb{E}_{\tilde{x}\sim\mathbb{P}_g} [D(\tilde{x})] - \mathbb{E}_{x\sim\mathbb{P}_r}[D(x)] + \\
        \lambda \mathbb{E}_{\hat{x} \sim \mathbb{P}_{\hat{x}}} [(||\nabla_{\hat{x}} D(\hat{x})||_2 - 1)^2 ]
\end{aligned}
\end{equation}
where, $\mathbb{P}_{\tilde{x}}$ is defined by sampling uniformly across straight lines between pairs of points sampled from $\mathbb{P}_r$ and $\mathbb{P}_g$.

\begin{figure*}[h]
\begin{center}
\includegraphics[width=13cm, height=7.5cm]{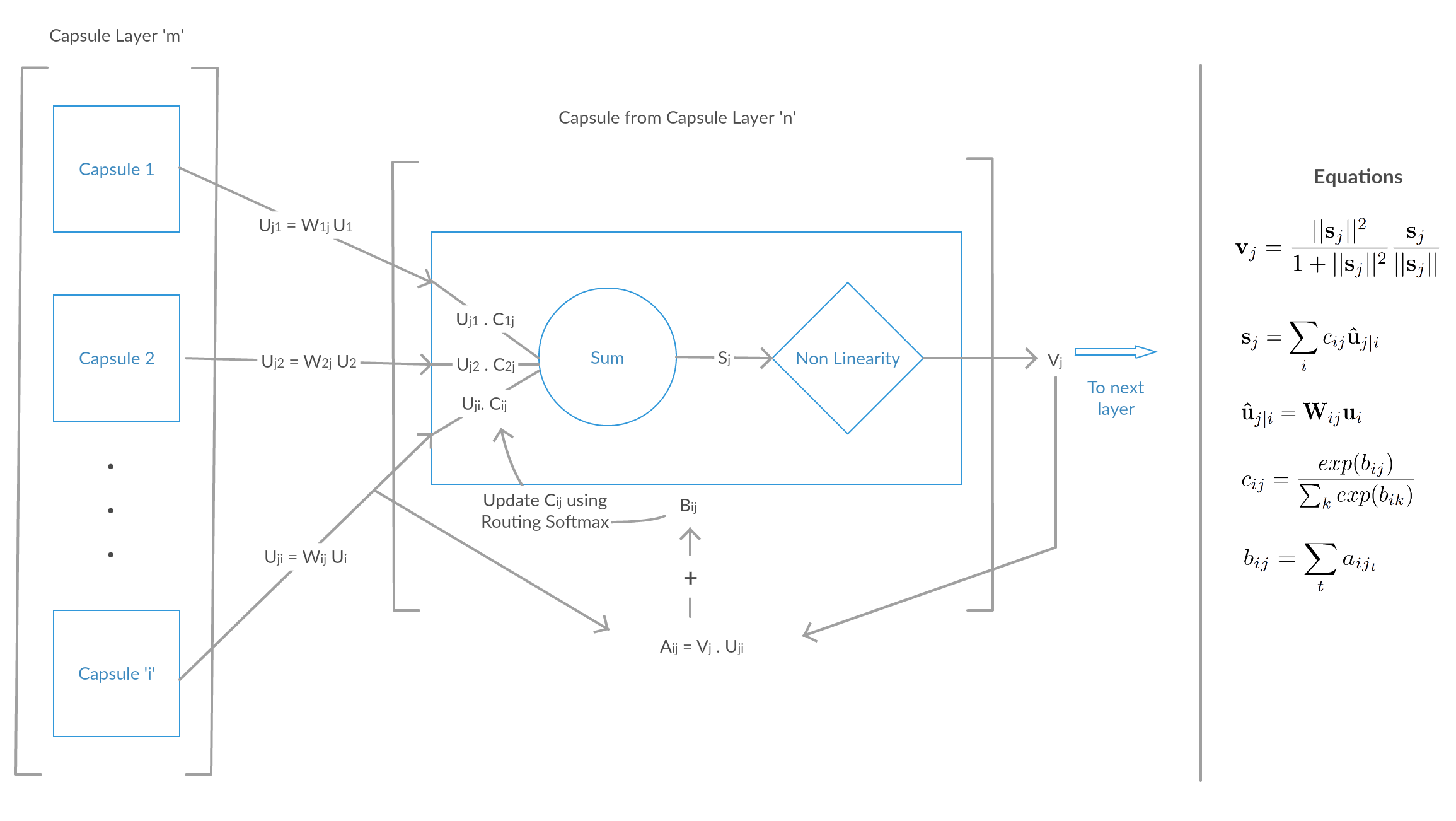}
\end{center}
   \caption{Dynamic Routing between Capsule Network Layers: The figure shows the attention like routing mechanism between capsule layers that allow a capsule to choose its parents via an iterative deterministic process.}
\label{RoutingArch}
\end{figure*}

While there are many variants of the basic GAN architecture, the core ideas have crystallized and recent innovations revolve around specializing GANs to aid in very specific applications. Deep Convolutional Neural Networks \cite{lecun1998gradient} have been the work-horses for the task of image synthesis for a while. DCGANS,  \cite{radford2015unsupervised} use CNNs as discriminators and Deconvolutional Neural Networks \cite{zeiler2010deconvolutional} as generators. CNNs capture localized features through the layers over varying granularity and use max-pooling to incorporate positional invariance of these features captured in an image. Despite providing the required variance in positions of the local features (in a limited manner), max-pooling leads to a form of lossy compression of the image features. Also, \cite{ruderman2018pooling} shows that the use of pooling helps CNNs only in the earlier epochs of the training and the CNNs that have been trained over greater number of epochs learn smoother filters that achieve the same performance as the CNNs with pooling layers that have been trained over similar number of epochs. Thus, rendering the employment of pooling layers unnecessary. CNN filters run convolutions over highly localized areas and through the deeper layers try to achieve positional invariance. Therefore,  CNNs end up learning limited spatial-relations between the features present in the image. This issue of a lossy feature learning was discussed by \cite{sabour2017dynamic}, which led to the introduction of Capsule Networks for image classification. 

\cite{sabour2017dynamic} developed the idea of Capsules, which are a group of neurons whose activity vector represents the instantiation parameters of a specific type of entity, such as an object, and the length of this activity vector represents the probability of the existence of the entity that the vector represents. Capsule Networks incorporate positional-equivariance as opposed to positional-invariance between the features of an image by using Dynamic Routing between Capsules. The attention-like Routing algorithm between layers allows Capsules from a given layer to learn the contributions from the relevant Capsules from the previous layer. This leads to Capsule Networks learning a richer representation of the features present in the images along with the relations between them on a more global scale.

Therefore, we explore Capsule Network, a more powerful network as a critic in a WGAN. A more powerful critic that can model the manifold better and reach optimality faster can provide better gradients for the generator to learn. Thus, helping the generator synthesize images of greater visual fidelity while seeing significantly lesser number of samples in comparison to WGANs that use a CNN critic. 
\newline
\newline
Following are the contributions of this paper:
\begin{enumerate}
  \item GAN architecture with a Capsule Network critic 
  \item A Split-Auxiliary critic architecture for using Capsule Networks for conditional image synthesis
  \item Quantitative analysis of the images synthesized by our proposed architectures
  \item Analysis of why Capsule GAN provides better coverage of the image space and diversity in the images synthesized   
\end{enumerate}

\section{Introduction to Capsule Networks} \label{CapsIntro}

\cite{sabour2017dynamic} developed Capsule Networks as parse trees carved out from a single multi-layer neural network where each layer is divided into many small groups of neurons called as Capsules, corresponding to each node in  the parse tree. Each Capsule vector represents the meta-properties of the feature/entity the Capsule represents and the overall length of the Capsule represents the probability of the presence of the entity the Capsule represents. Each active Capsule chooses its parents from the layer above it using an iterative attention-like routing process. This dynamic routing process replaces the max-pooling step from CNNs allowing for better feature globalization and smarter feature compression.

Since the length of each capsule represents the probability of the presence of the entity it represents, a non-linear "squashing" function is used to shrink the length of the vector $s_j$ between 0 and 1, which is denoted by the vector $v_j$, the output of the capsule $j$. 

During the routing process from capsule $i$ of a given layer to capsule $j$ of the next layer, the output of capsule $i$, $u_i$ is first multiplied by the matrix $W_{ij}$ to give $u_{ji}$. $s_j$ is then calculated as the weighted sum over all $u_{ji}$ coming from the previous layer to capsule $j$, weighted over the coupling coefficient, $c_{ij}$. The coupling coefficient $c_{ij}$ is calculated as the softmax over all $b_{ij}$, which is the summation of the agreements, $a_{ij}$ between the individual input capsules and output of capsule $j$ over all the iterations. The agreement, $a_{ij}$, is calculated as the dot product of the output, $v_j$ and the incoming vector,  $u_{ji}$.

The Capsule Network introduced by \cite{sabour2017dynamic} used a marginal loss over the capsules in the final layer for optimization along with a reconstruction loss which helps in regularization. However, the architectures proposed in this paper do not incorporate the reconstruction loss. The marginal loss, $L_k$ is defined as following,
\begin{equation} \label{marginalLoss}
\begin{aligned}
L_k = T_k(max(0,m^+ - ||v_k||^2))^2 + \\
\lambda(1-T_k)(max(0, ||v||^2 - m^-))^2    
\end{aligned}
\end{equation}

where $m^+ = 0.9$, $m^- = 0.1$, $T_k = 1$ iff the entity of class $k$ is present else, $T_k = 0$. $\lambda = 0.5$ is used as a a down-weighting factor to prevent shrinking of activity vectors in the early stages of training.

\begin{table}[h]\centering
 \label{trainingPerf}
\begin{tabular}{lcccc}
\toprule
\multirow{2}{*}{Epochs} & \multicolumn{2}{c}{MNIST} & \multicolumn{2}{c}{Fashion-MNIST} \\
\cmidrule(lr){2-3} \cmidrule(lr){4-5}
 & CNN & CapsNet & CNN & CapsNet \\
\midrule
1 & 91.09 & 98.51 & 48.72 & 84.25 \\
2 & 93.53 & 92.22 & 73.99 & 86.53 \\
3 & 95.04 & 99.41 & 75.36 & 87.91 \\
4 & 96.30 & 99.58 & 78.64 & 88.97 \\
5 & 97.17 & 99.63 & 81.02 & 90.05 \\
\hline
\end{tabular}
\newline
\caption{Capsule Network v/s Convolutional Neural Network test accuracy comparison(\%)}
\end{table}

\begin{figure*}[!ht]
  \begin{center}
  \includegraphics[width=13cm, height=7cm]{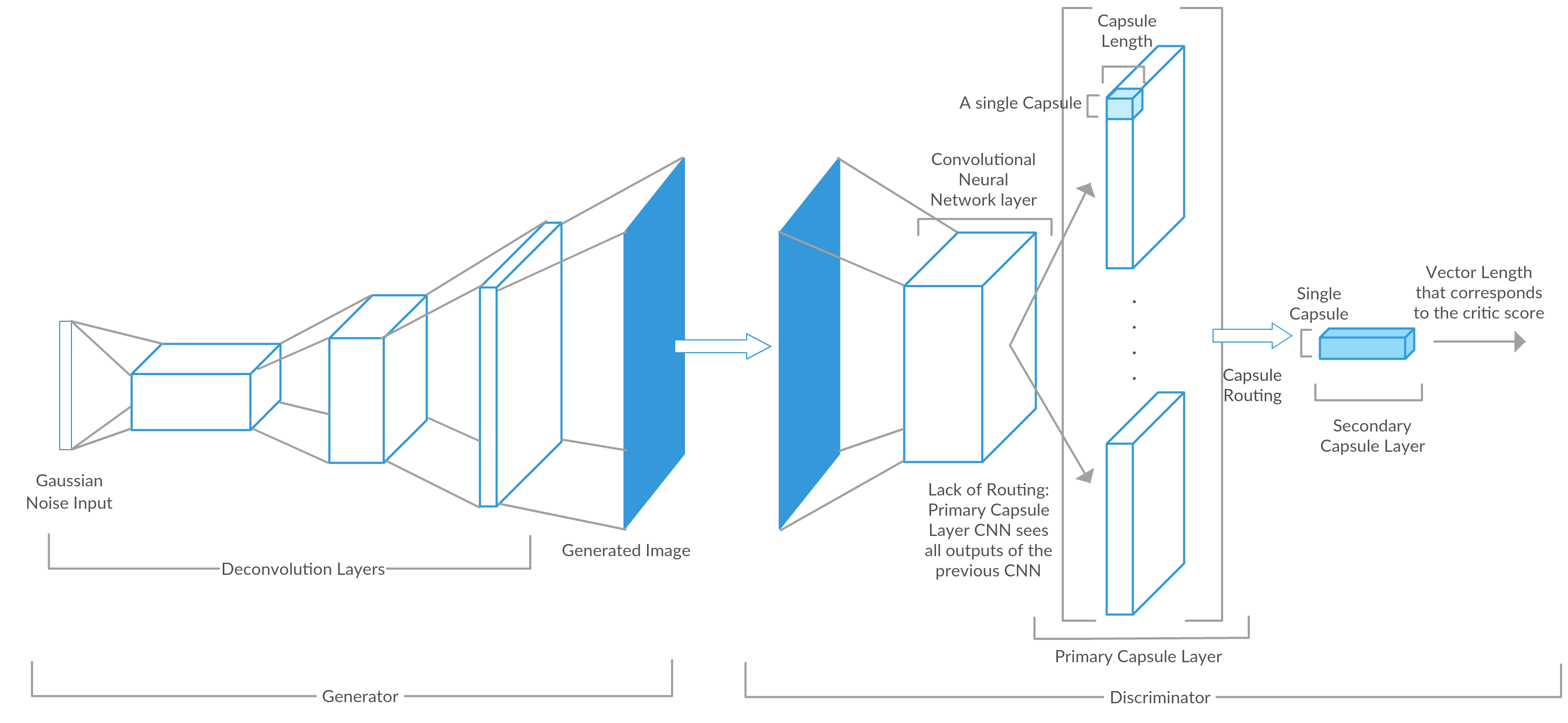}
  \end{center}
  \caption{Capsule GAN Architecture}
  \label{generalArch}
\end{figure*}

\section{Key ideas}\label{keyIdeas}

\subsection{Equivariance is better than Invariance}

Capsule networks are able to learn the features and the relations between them better than Convolutional Neural Networks because the Primary Capsule layer looks at all the the features of the input image and the routing process determines the set of global features that contribute to a capsule in the Secondary Capsule layer. Capsule Networks are based on the principal of positional equivariance whereas, CNNs are based on the principal of positional invariance of the features. Convolutional Neural Networks lose the spatial relationships between features through successive layers especially if using max-pooling as it brings greater positional-invariance and proves as a lossy form of feature compression.
Also, the fact that each Capsule encodes the properties of the entity/feature it represents, enables Capsule Networks to model the distribution in greater granularity, making them more robust to small affine transformations \cite{sabour2017dynamic}.

\subsection{Better the critic, better the generator}

The concept of using a Wasserstein-1 distance as a loss metric \cite{salimans2016improved} requires replacing a discriminator, that outputs the probability of an image being real or fake with a critic that assigns a high score to a real image. The critic is trained to maximize the Wasserstein-1 distance between the scores assigned to the real and fake images whereas the generator is trained to increase the score that the critic churns out for the images it synthesized. The true gradients being more meaningful, guide the generator to synthesize more realistic images consistently. The better the gradients, the better the generator learns but for the gradients to be of healthy, the critic must reach optimality. Therefore, the faster the critic reaches the optimality, the faster the generator learns. The quality of the gradients also depends on the capability of the critic to learn the image manifold since, a poorly learned distribution wont give accurate scores to the images. Therefore, the better the critic learns the manifold, the better the generator gets.

Referring to the previous section, we see that the principal of feature equivariance helps Capsule Networks perform better than the CNNs modelled after feature invariance. The improvement in performance comes in the form of learning the distribution faster and better. Therefore, replacing the CNN critic with a Capsule Network should present us with improvements in the the quality and reduction in the amount of data required to synthesize images with visual fidelity.

\subsection{Critiquing and classification are supplementary}

Critiquing requires the critic to learn the distribution of the real images and generate high scores to the images that belong to them. Whereas, classification requires that the classifier learn the key features occurring in the dataset of images and use them to classify the the images according to the features present in them. As described in \cite{odena2016conditional}, the using class information for training a GAN aids the GAN to model the structure better. Being able to learn features that play a key role in discriminative tasks can help the critic learn more about the distribution. Capsule Networks have achieved state-of-the-art performance for classification. This throws light on the fact that Capsule Networks can capture features that are better than captured by vanilla CNNs, that are key in classification. This points into a direction where Capsule Networks can be used to extract discriminative features that can help in critiquing. Instead of of using an ensemble of a generator and a classifier, we use a split-auxiliary architecture, where the network remains same for the critique as well as the classifier up till the last layer and the penultimate layer feeds its features to two different layers with different purposes.
Capsule Networks have achieved state-of-the-art performance for classification. This helps the Primary Capsule layer learn the features necessary for discrimination while also building structure in features to help the critic score the samples.

\section{Architectures}

In this section, we describe the architectures that this paper proposes for random as well as conditional image synthesis.

\subsection{Capsule GAN}\label{DCapsArch}

This architecture uses Capsule Networks as a discriminator in place of a Convolutional Neural Network used in DCGANs. It can be seen in Fig. [\ref{generalArch}], our Capsule Network uses two Capsule layers: Primary and Secondary Capsule layers, in which, there is no routing between the Convolutional layer and the Primary Capsules. Routing exists only between the Primary and Secondary Capsules. The Secondary Capsule layer consists of only one Capsule and the "squashing" non-linearity used in the Primary Capsule layer is not applied here. The length of the activity vectors of this Capsule represents the score of the critic. We use the critic loss function described in Eqn. [\ref{GPeqn}], where the critic $D$ returns the length of the output capsule. Whereas, the generator loss function is described as follows:

\begin{equation} \label{generalGLoss}
L_G = -\mathbb{E}_{x\sim P(z)} -D(G_\theta (x))
\end{equation}

where  $P(z)$ represents the prior distribution serving as the input to the generator $G_\theta$. 

\subsection{Conditional Split-Auxiliary Capsule GAN}\label{CCapsArch}

For conditional generation, apart from receiving random variables generated from a Gaussian distribution, the generator also receives the class from which it must synthesize the image. The two vectors are then concatenated and utilized by the generator as a latent space representation of the image to be synthesized.   

The discriminator uses a variation of the auxiliary-conditional architecture described by \cite{odena2016conditional}. The discriminator consists of a similar architecture up to the Primary Capsule layer as described in Section \ref{DCapsArch}. The Primary Capsules then serve as an input for two different Secondary Capsule layers: Primary Critic and Secondary Classifier. The Primary Critic scores the input images as being fake or real whereas,  the Secondary Classifier classifies the image into the class label it belongs to. The Wasserstein Loss from the Primary Critic and the Marginal Loss from the Secondary Classifier (Eqn. [\ref{marginalLoss}]) are then coupled together. 

\begin{figure*}[h]
    \begin{center}
    \includegraphics[width=12cm, height=6cm]{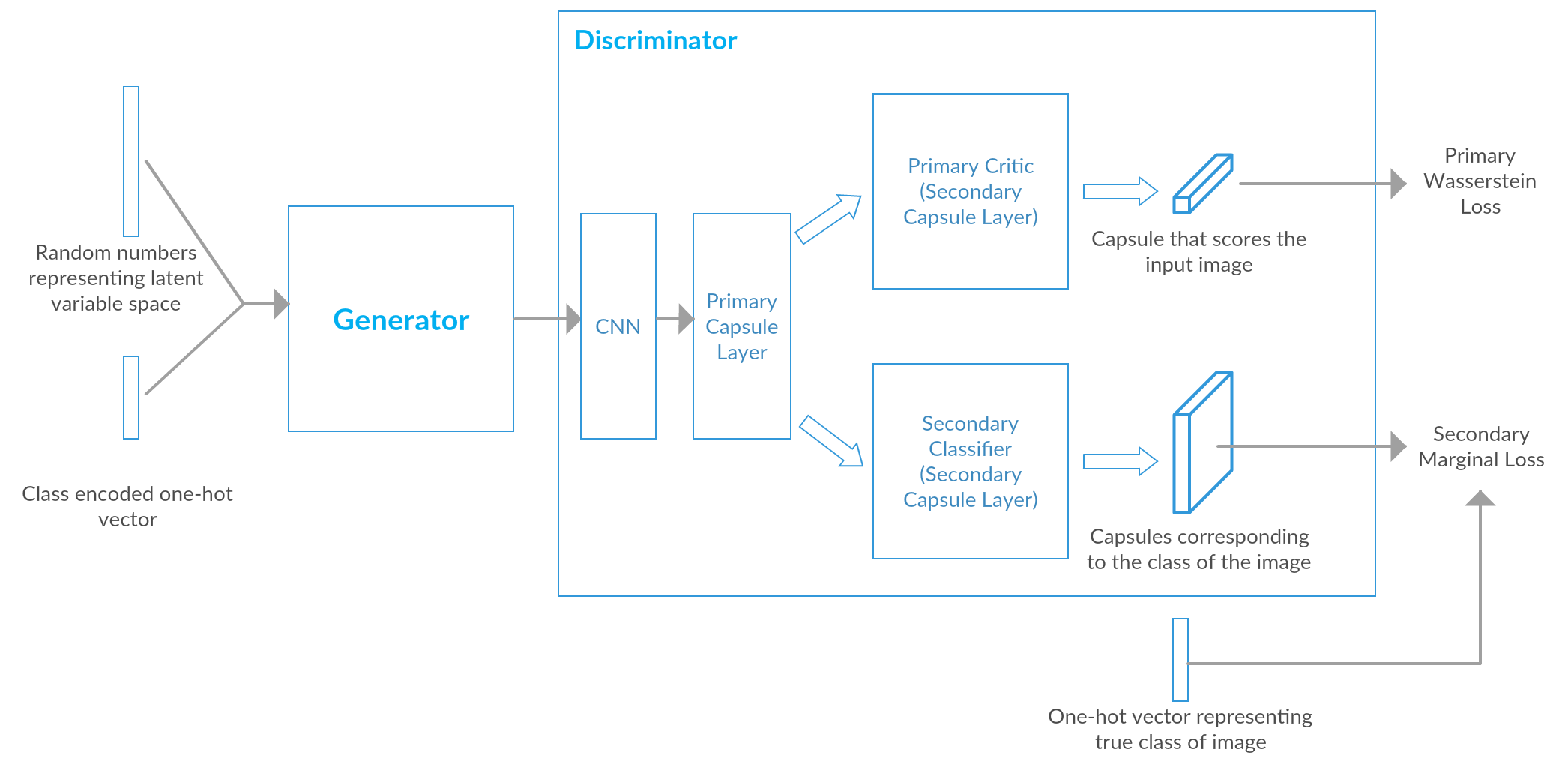}
    \caption{Architecture of Conditional Split-Auxiliary Capsule GAN}
    \end{center}
    \label{condArch}
\end{figure*}

Let the Primary Critic Loss be $L_{P}$ (Eqn. [\ref{GPeqn}]), Secondary Marginal Loss be $L_{S_c}$ (Eqn. [\ref{marginalLoss}]) and the Generator Loss from Eqn. [\ref{generalGLoss}] be denoted by $L_{G_{W}}$. The losses for the Discriminator($L_D$) and the Generator($L_G$) is given as follows:
\begin{equation}\label{condDLoss}
L_D = L_{P} + L_{S_{x \in P(r) | y = c}}(x) + L_{S_{x \in P(z) | y = c}}(G_\theta (x))    
\end{equation}
\begin{equation}\label{condGLoss}
L_G = -L_{G_{W}} + L_{S_{x \in P(z) | y = c}}(G_\theta (x))    
\end{equation}

where $G_\theta$ is the generator with parameters $\theta$, $P(r)$ corresponds to the probability distribution of the dataset, $P(z)$ corresponds to the probability distribution of the prior to the generator, $k\in \{P(r),P(z)\}$, $y$ is the class label of the image, and $c$ corresponds to the intended class of the image. The Secondary Marginal Losses force the discriminator to the learn the representation of an image conditionally over a label. The split architecture allows for the Primary and Secondary Capsule Classifiers to borrow from the same set of extracted features in the Primary Capsules for 2 tasks - critiquing the validity of the image and classifying the class of the image. Apart from helping the Primary Capsules learn features for the class of an object, the split-architecture also helps reduce computational overheads due to a completely autonomous second classifier. 

\section{Results}

\subsection{Images}

\begin{figure}[b]\label{nonCondIm}
\begin{center}
\subfloat[]{\label{nonCondIma}{\includegraphics[width=0.2\textwidth]{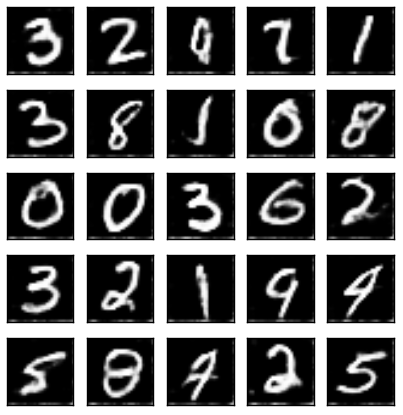}}}\hfill
\subfloat[]{\label{nonCondImb}{\includegraphics[width=0.2\textwidth]{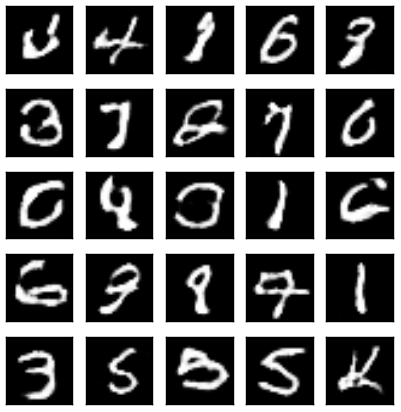}}}
\caption{Fig.(a) is generated by our architecture whereas, Fig.(b) is generated by Improved Wasserstein GAN, both trained over 5 epochs.}
\end{center}

\end{figure}

\begin{figure}[b]\label{nonCondFashIm}
\begin{center}
\subfloat[]{\label{nonCondImc}{\includegraphics[width=0.2\textwidth]{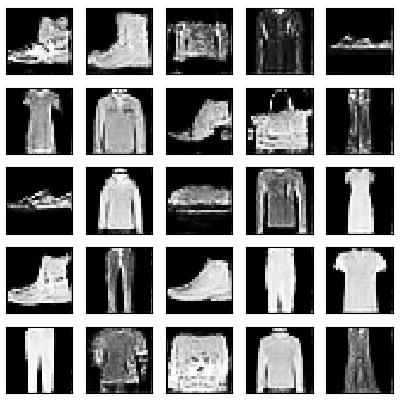}}}\hfill
\subfloat[]{\label{nonCondImd}{\includegraphics[width=0.2\textwidth]{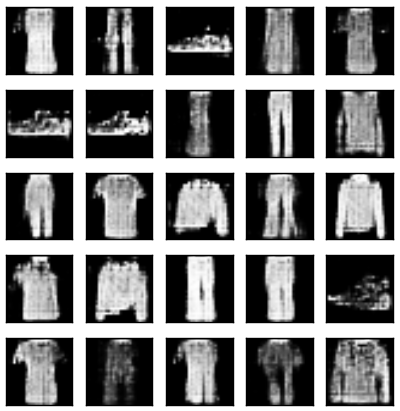}}}
\caption{Fig.(a) is generated by our architecture whereas, Fig.(b) is generated by Improved Wasserstein GAN, both trained over 5 epochs.}
\end{center}

\end{figure}

\begin{figure*}[!h]

\subfloat[]{\label{CondE100a}{\includegraphics[width=0.4\textwidth]{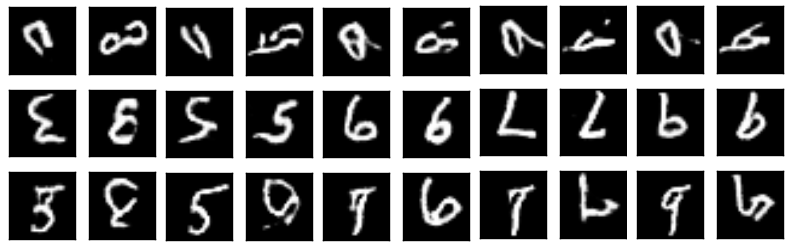}}}\hfill
\subfloat[]{\label{CondE100b}{\includegraphics[width=0.4\textwidth]{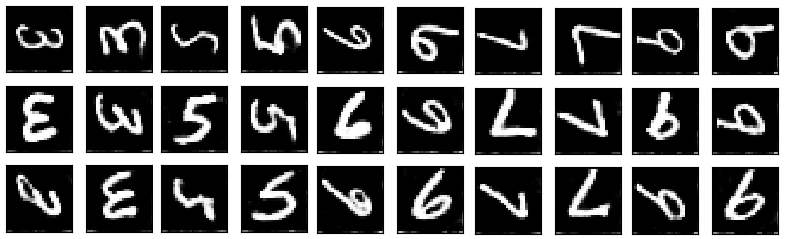}}}
\caption{Images generated by \textbf{(a) }Conditional Improved Wasserstein DCGAN trained over MNIST for 100 epochs \textbf{(b)} Split-auxiliary Conditional Capsule GAN trained over MNIST for 5 epochs. The digits being compared are - 3, 5, 6, 7, 9. We can can observe that the images generated by our model are visually much better than the images generated by DCGAN despite being trained over significantly lesser number of samples.}
\label{CondE100}

\end{figure*}

\begin{figure*}[h]
    \begin{center}
    \includegraphics[width=0.8\textwidth]{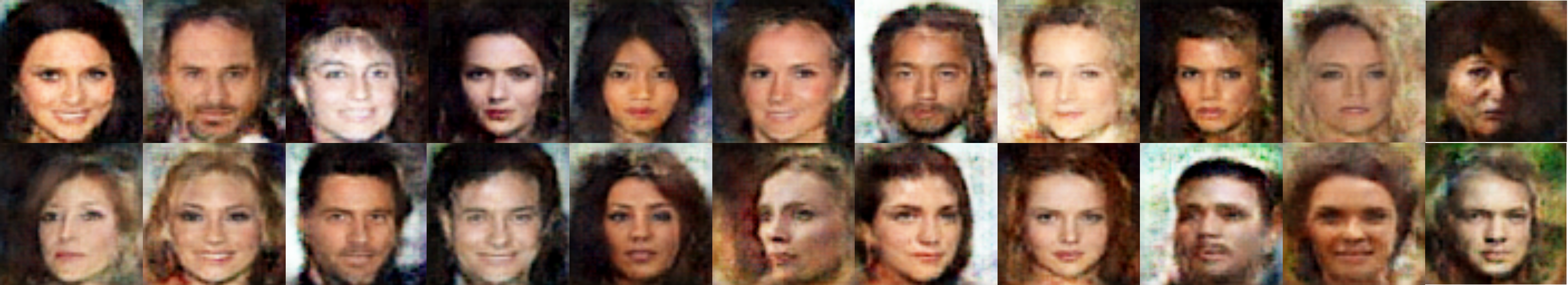}  
    \end{center}
    \caption{64x64 resolution images generated from CelebA in 50 epochs by Capsule GAN}
    \label{celebA}
\end{figure*}

As a part of the assessment, we have trained out architectures on multiple datasets. We have compared the results from our proposed architectures with the images generated by Improved Wasserstein GAN, that utilizes a CNN critic with a similar number of backpropagation trainable parameters. We can see in Fig. [\ref{nonCondIma}] and Fig. [\ref{nonCondImc}] that the proposed unconditional architecture, trained on  MNIST \cite{lecun1998mnist} and Fashion-MNIST \cite{xiao2017fashion} datasets, synthesizes images with high visual fidelity even in the earlier epochs. We have also trained our model on CelebA \cite{yang2015facial} to synthesis images with a resolution of 64x64. The results in Fig. [\ref{celebA}] have been generated in 50 epochs.

\begin{figure}[h]
    \begin{center}
    \includegraphics[width=0.5\textwidth]{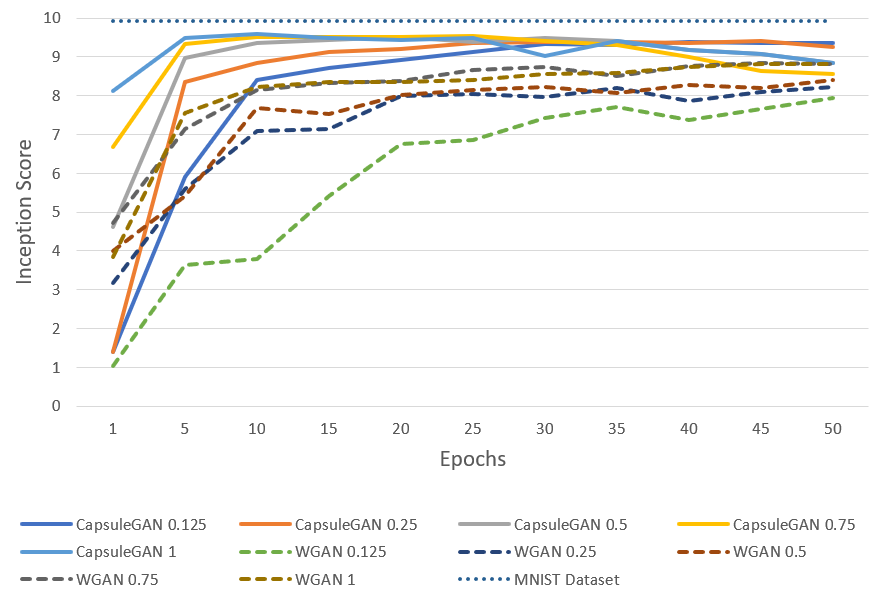}    
    \end{center}
    \caption{Inception Scores of Capsule GAN and IWGAN being trained over a fraction of the MNIST dataset}
    \label{MNISTIS}
\end{figure}

We also synthesized rotated MNIST images by training Conditional Improved Wasserstein GAN and Conditional Split-Auxiliary Capsule GAN on an MNIST dataset which has images with rotations of: 0, 90, 180, 270 degrees. We can see in Fig. [\ref{CondE100b}], that despite our architecture being trained for only 5 epochs, the quality of the images generated by it surpasses the quality of the images generated by Conditional Improved Wasserstein GAN which was trained over 100 epochs over the same dataset. One can clearly see that the our model has been able to pick up really strong features even in the earlier epochs to distinguish between a '6' and a rotated '9'. It is easy to see that our model has learned to distinguish between the two by the curvature of the tail in the two digits. These results bolster our key idea of having a split-classifier that optimizes the class conditionals and the process of critiquing simultaneously in one network. 

\subsection{Evaluation}

\begin{table*}[h]
\begin{center}
 
\begin{tabular}{lccccccccccc}

\toprule
Dataset & Model & 0 & 1 & 2 & 3 & 4 & 5 & 6 & 7 & 8 & 9 \\
\midrule
\multirow{4}{*}{MNIST} & {\small IW DCGAN} & 2.07 & 1.74 & 3.00 & 2.24 & 3.17 & 3.59 & 2.29 & 1.25 & 3.56 & 2.61 \\
                       & Capsule WGAN (Ours) & $\bold{0.30}$ & $\bold{0.16}$ & $\bold{0.46}$ & $\bold{0.15}$ & $\bold{0.26}$ & $\bold{0.48}$ & $\bold{0.42}$ & $\bold{0.15}$ & $\bold{0.41}$ & $\bold{0.35}$ \\
                       & Cond. IW DCGAN & 0.59 & 0.23 & 0.97 & 1.777 & 0.25 & 1.559 & 0.60 & 0.51 & 1.372 & 0.96 \\
                       & Cond. SAC GAN (Ours) & $\bold{0.13}$ & $\bold{0.15}$ & $\bold{0.13}$ & $\bold{0.17}$ & $\bold{0.20}$ & $\bold{0.13}$ & $\bold{0.12}$ & $\bold{0.24}$ & $\bold{0.15}$ & $\bold{0.16}$ \\
\hline
\multirow{2}{*}{CelebA} & IW GAN & 0.73 & 0.94 & 0.46 & 1.66 & $\bold{1.345}$ & 0.41 & 1.53 & 1.953 & 2.82 & 0.56\\
                        & Capsule WGAN (Ours) & $\bold{0.42}$ & $\bold{0.79}$ & $\bold{0.44}$ & $\bold{1.28}$ & 2.16 & $\bold{0.24}$ & $\bold{1.19}$ & $\bold{1.93}$ & $\bold{0.73}$ & $\bold{0.31}$\\
\hline
\end{tabular}
\caption{FID Scores ($\times10^{-2}$) for GANs trained on MNIST and CelebA datasets over classes 0 - 9. Lower is better.}
\label{FID}
\end{center}
\end{table*}

Evaluating the images synthesized by generative model is a non-trivial task. One intuitive way to judge the images qualitatively is by the use of human annotators but it is a highly subjective process and the results vary greatly, even the ones coming from single person. Therefore, Inception Score, a quantitative measure, was introduced by \cite{salimans2016improved}, which is given by $exp(\mathbb{E}_{x}KL(p(y|x)||p(y))$. Inception score is found to be correlated with the human judgment of image quality and therefore, is one of the most widely used metric for evaluating image generative systems. 

The images synthesized by our architectures achieve state-of-the-art results for MNIST dataset and that too with significantly lesser amount of training data as well as training epochs. Referencing Fig. [\ref{MNISTIS}], we can see that the  Capsule GAN architectures achieve very high Inception Scores from the early training epochs itself. The ability to achieve such high scores in by looking at just $1/8th$ of the samples is a testament of the ability of Capsule Networks to encode a dataset's distribution in a much better manner. The top Inception Score achieved by our architecture is 9.58, whereas the test partition from MNIST achieved a score of 9.92. The conditional architecture were also able to achieve state-of-the-art performance with a score of 9.99, achieving almost the theoretical limit for equal distribution of perfect samples from each class. 

However, as many have pointed out \cite{barratt2018note, borji2018pros, heusel2017gans}, Inception Score does not take diversity of the images within a class into consideration. The generator can get away with a high score even if it replicated one image per class. Therefore, \cite{gurumurthy2017deligan} tried to introduce the modified-Inception Score, given by $e^{\mathbb{E}_{x_i} [\mathbb{E}_{x_j}[KL(p(y|x_{i})||p(y|x_{j})) ] ]}$, to reward the high entropy of class-conditional probabilities of images within a class. But this doesn't necessarily mean all the images in the given will be of good quality. This score would fail in a scenario where the generator has synthesized diverse, yet perfect images within a class and for all the classes. To address problems arising out of evaluating generated images over void, \cite{heusel2017gans} introduced the Frechet Inception Distance Score that uses the Wasserstein-2 distance between the activations for the real and generated images. It assumes that the activations from the two datasets follow a Gaussian distribution given by $(\mu_r,C_r)$ and $(\mu_f,C_f)$ and the lesser the distance between them, the better are the samples generated. It is given as follows,
\begin{equation} 
    ||\mu_r - \mu_f||^{2}_{2} + Tr(C_r + C_f -2(C_{r}C_{f})^{1/2}
\end{equation}

We calculate the FID Score for the generated images using a trained Capsule Network classifier. We use a Capsule Network because it achieves state of the art performance on the datasets with faster training time. With Inception Network, we were able to achieve only 98.3\% accuracy on MNIST, whereas we were able to get 99.72\% accuracy with Capsule Networks. Since Capsule Network can capture better features than the Inception Network, it will be a better judge of the features present in the synthesized images. 

For MNIST, where use the activations of the Secondary Capsule whose length is the maximum. Whereas, for CelebA, there can be multiple classes present in an image and therefore, we consider a class to be active if the vector length corresponding to its Secondary Capsule is greater than 0.5. The Secondary Capsules of active classes are then stacked class-wise for calculating the FID. We show the class-wise FID scores for all the classes of MNIST and the first 10 classes for CelebA. Referencing Table [\ref{FID}], we can see that our model achieves significantly better FID scores on MNIST as well as CelebA.

\section{Image Diversity}

In the earlier sections, we have discussed about Capsule Networks being able to capture spatial relationships better and therefore, better features in images when compared to CNNs. A direct impact was seen in the quantity of the training data and epochs required to achieve state of the art results. The Primary Capsule layer looks at the complete image and all the Capsule contribute to the Capsules in the Secondary Capsule layer weighted by the agreements between them. This gives a better global view of the features when compared to CNNs which only look at local features progressively through the layers. Thus, Capsules are able to capture certain features which are completely missed out by the CNNs. Apart from having an impact on the training statistics, there is a strong impact on the diversity of the images generated by a GAN having a CNN discriminator. Referring to one of our key ideas about Wasserstein GAN critics that the generators image synthesis quality is only as good as the critic that judges it. If the critic is unable to capture all the aspects of the true image distribution, then the Wasserstein Loss tries to pull the distribution of the synthesized images to the more occluded critic's understanding of the distribution. Thus, limiting the overall coverage of the generated distribution over the actual distribution.

\begin{figure}[t]
    \begin{center}
    \includegraphics[width=0.45\textwidth]{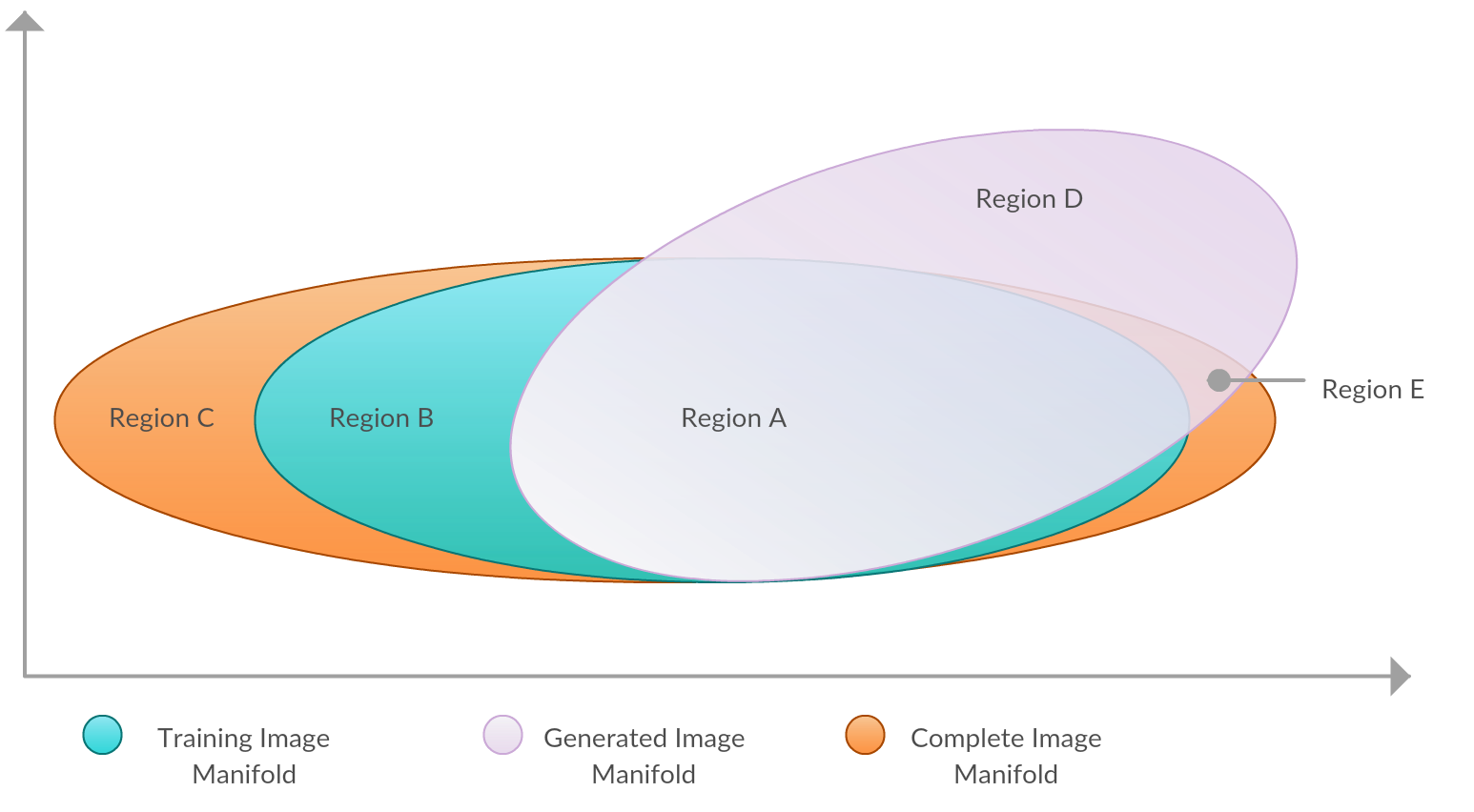}
    \caption{Overlap of complete image manifold, training data manifold and generative model generator model learned manifold}
    \label{ManifoldSimulation}
    \end{center}
\end{figure}

In Fig [\ref{ManifoldSimulation}], there are 3 different manifolds - training image manifold, manifold captured by the generator and the complete image manifold. It is possible that the training images themselves may not be able to capture the entire manifold of the possible images, therefore, in this case, the training image manifold is a smaller subset of the possible manifold. Elaborating on Fig. [\ref{ManifoldSimulation}], $A$ represents the region of the training images covered by the generator, whereas $B$ represents the region of training images missed by the generator which corresponds to lack of coverage. Region $C$ represents the manifold covered neither by the training images, nor the generator whereas, $E$ represents the region of true novelty, where the generator is producing images that do not belong to the manifold of the training samples but are still visually correct and region $D$ represents the region of incorrect samples being generated by the generator. While optimizing the GAN loss function, one would want that the GAN generator is able to capture the entire manifold of the training images, thus, leading to coverage as well as diversity. 
\begin{figure}[t]
    \begin{center}
    \includegraphics[width=0.45\textwidth]{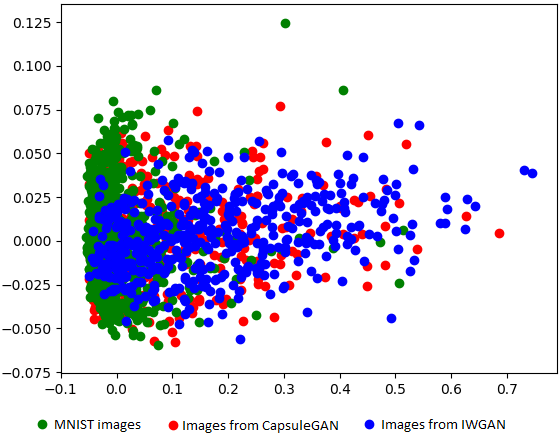}
    \caption{Projection of Capsule representation of MNIST class-5 images in $\mathbb{R}^2$ by projecting Capsule vectors on the two largest PCA componenets of Capsules corresponding to the training images}
    \label{ImageProjection}
    \end{center}
\end{figure}
Upon visualizing the Capsule representations of the MNIST images in Fig. [\ref{ImageProjection}],  we can see that the images generated by Capsule GAN has a greater coverage over the secondary principal axis in comparison to that of the images generated by the IWGAN. Referencing to Fig. [\ref{ManifoldSimulation}], we can see that the overlap region, $A$, for Capsule GAN is much larger than the overlap region for IWGAN. Most of the images generated by IWGAN are packed closely and have lesser coverage over the real data manifold. The point to note is that despite achieving visual fidelity in images, IWGAN was not successful in capturing the complete manifold discovered by the Capsule Network projection, in other words, it had inferior coverage as well as diversity in the attributes of the samples it generated. Since, these are projections from the Capsule space, we attribute such a behaviour of the IWGAN to the lack of ability of the CNN critic to learn the features unearthed by the Capsule Network. The principal of serializing positional invariant layers leads to a systematic failure of coverage, making the IWGAN oblivious to the features that Capsule GAN is able to pick up.

\section{Conclusion and Future Work}
Building up on the foundation that positional-equivariance is superior to positional invariance, successive convolutional layers, especially with the use of intermittent pooling layers, leads 
to a lossy compression of the image. on the other hand, Capsule Networks, built on the principal of positional-equivariance, look at the the whole image at once and map the important features from one layer to the other using agreements via Dynamic Routing.
Thus, making them better feature learners which enables them to get better performance in comparison to CNNs, even with significantly lesser amount of training data and training epochs. 

Upon exploring the application of a Capsule Networks as a critic based on the ideas a) the faster the critic reaches optimality, the faster the generator learns to produce better images and b) the generator
is limited by its critic's capabilities, we find that indeed, IWGANs with Capsule Network as a critic produce images with visual fidelity much faster when compared to IWGANs that use CNNs with a similar number 
of training parameters. We also found that since Capsule Networks are better feature encoders, it is still able to perform much better with lesser training data.

We also explored the coverage and diversity of the images synthesized by Capsule GAN to be greater than that of the images produced by the IWGAN. Successive positional-invariant convolution layers become blind to certain key features that are captured by the Capsule Network. This leads to the CNN critic to learn a limited view of the manifold, thus, providing gradients to generator limited to its understanding.

The fact that replacing Convolutional Neural Networks with Capsule Networks could bring significant improvements to the overall performance of GANs, points us in a direction which encourages exploration in the application of the concept of capsules to the generator. This would require the use of inverting the Capsule Network to generate an image encoded in the Secondary Capsule representation of the image. However, inverting the Dynamic Routing process is non-trivial. To overcome this one might look into the Capsule Networks that utilize Expectation Maximization Routing where, each of the Capsules in a given layer represent data points that belong to distributions represented by Capsules in the next layer.

{\small
\bibliographystyle{ieee}
\bibliography{egbib}
}

\end{document}